\documentclass[letterpaper]{article} 
\usepackage{aaai24}  
\usepackage{times}  
\usepackage{helvet}  
\usepackage{courier}  
\usepackage[hyphens]{url}  
\usepackage{graphicx} 
\usepackage{natbib}  
\usepackage{caption} 
\usepackage{algorithm}
\usepackage{algorithmic}
\usepackage{color, soul}
\usepackage{subfigure}
\usepackage[utf8]{inputenc}
\usepackage{amsmath,amsthm}
\usepackage{amssymb}
\usepackage{mathtools}
\usepackage{multirow}

\usepackage{newfloat}
\usepackage{listings}
\usepackage{newfloat}
\usepackage{listings}

\theoremstyle{definition}
\newtheorem{definition}{Definition}

\DeclareCaptionStyle{ruled}{labelfont=normalfont,labelsep=colon,strut=off} 
\floatstyle{ruled}
\newfloat{listing}{tb}{lst}{}
\floatname{listing}{Listing}
\title{A Formalism and Approach for Improving Robustness of Large Language Models Using Risk-Adjusted Confidence Scores}
\author{
    Ke Shen, Mayank Kejriwal
}
\affiliations{
    \textsuperscript{\rm 1}Information Sciences Institute \\

    University of Southern California \\
    Marina del Rey, California, United States 90292\\
    keshen, kejriwal@isi.edu
}

\usepackage{bibentry}

\begin{document}

\maketitle

\begin{abstract}
Large Language Models (LLMs), such as ChatGPT, have achieved impressive milestones in natural language processing (NLP). Despite their impressive performance, the models are known to pose important risks. As these models are deployed in real-world applications, a systematic understanding of different risks posed by these models on tasks such as natural language inference (NLI), is much needed. In this paper, we define and formalize two distinct types of risk: decision risk and composite risk. We also propose a risk-centric evaluation framework, and four novel metrics, for assessing LLMs on these risks in both in-domain and out-of-domain settings. Finally, we propose a risk-adjusted calibration method called DwD for helping LLMs minimize these risks in an overall NLI architecture. Detailed experiments, using four NLI benchmarks, three baselines and two LLMs, including ChatGPT, show both the practical utility of the evaluation framework, and the efficacy of DwD in reducing decision and composite risk. For instance, when using DwD, an underlying LLM is able to address an extra 20.1\% of low-risk inference tasks (but which the LLM erroneously deems high-risk without risk adjustment) and skip a further 19.8\% of high-risk tasks, which \emph{would} have been answered incorrectly.


\end{abstract}
\section{Introduction}
The advent of powerful transformer-based discriminative language models and more recently, generative \emph{large language models} (LLMs) has yielded significant improvements even on difficult natural language inference (NLI) tasks \cite{wang2022pre, sun2022survey, li2022language, zhou2023comprehensive, treviso2023efficient}. However, problems with such models, including hallucination, bias, and over-confidence, have led to renewed concerns about understanding (and minimizing) the risks posed by these models \cite{hendrycks2021many, sai2022survey, ji2023survey,zhuo2023exploring, ferrara2023should}. Minimization of risk can not only lead to greater robustness in these models, but also allow us to understand the properties of the models themselves better. More practically, a clear understanding of risk is necessary before these models can be trusted and deployed in critical real-world situations. 

Traditionally, in machine learning \cite{murphy2012machine}, the risk of a model's inference tended to be directly equated to its confidence score: lower confidence was assumed to signal increased risk (at least from the model's own perspective), implicitly defined as the probability that the prediction made by the model is correct. Even if the assumption were true (and much work in deep learning has showed that it is not), this one notion of risk is incomplete. It does not capture the inherent risk in (for example) an ambiguous inference task with no, or a highly controversial, correct answer in the ground truth. More interesting notions of risk also arise if the model is given a choice to \emph{abstain} from making a prediction. Intuitively, given such an ability, if the model is too aggressive in making predictions, it is taking on much more risk than is appropriate (given its underlying accuracy and performance characteristics), whereas too much caution results in a suboptimal tradeoff between risk and reward.

Once defined and identified, such risks need to be minimized using systematic approaches. In early literature on deep neural networks (DNNs),  one option was to use the raw confidence of the DNN, usually obtained from the final softmax output layer, to make decisions on abstaining or otherwise trusting the output \cite{vasudevan2019towards, brown2020language}. However, it was quickly recognized that this measure of confidence actually results in over-confidence, and may not be aligned with the genuine confidence of the model in its own outputs \cite{chorowski2017towards, zhou2023navigating}. Similar problems, in addition to others, such as hallucination and bias, have also been noted and discussed for LLMs.

This paper aims to move beyond a single and often implicit definition of risk by introducing a risk-centric framework that defines two different types of risk, and proposing metrics for evaluating LLMs on these two risks. We also propose a risk-adjusted calibration method for minimizing some of these risks.  Specific contributions are enumerated below:

\begin{itemize}
    \item We formalize and present a risk-centric evaluation framework for language-based inference models (such as LLMs), by identifying and defining two types of risks (decision and composite risk), and four novel metrics to measure these risks. To ensure that the framework captures different levels of robustness, we present both in-domain and out-of-domain evaluations for decision risk. 
    
    \item We present a novel risk-adjusted calibrator (called DwD) for adjusting the raw confidence of an underlying LLM to navigate decision and composite risks better. DwD is compatible with both discriminative and generative LLMs, and is designed to be general by being only loosely coupled to the underlying LLM. While it uses the LLM's raw confidence scores as features, it also relies on a robust set of features designed to capture as much available information about the prompt as possible. 
    
    \item We conduct a detailed set of systematic experiments, using both a RoBERTa-based model and ChatGPT, against three decision rule baselines on four NLI benchmarks. Our results demonstrate the practical utility of the proposed framework, as well as the competitive performance of DwD for minimizing both types of risk. DwD reduces in-domain and out-of-domain decision risks more for RoBERTa by margins of 28.3\% and 18.9\%, respectively. Similarly, using a risk sensitivity metric, DwD reduces additional composite risk of ChatGPT NLI by a margin of 17.9\%, compared to the next best baseline.

\end{itemize}

\section{Related Work}\label{related_work}
Discriminative and generative large language models (LLMs) have recently achieved impressive performance on multiple inference tasks \cite{bhargava2022commonsense, zhao2023survey}. However, due to documented problems such as hallucination and bias, the research focus is shifting from merely quantifying accuracy to an in-depth, context-sensitive probing of LLMs' robustness, particularly when confronted with risky or uncertain situations \cite{wang2023robustness,shen2023experimental,liang2022holistic}.

A pivotal aspect of understanding the robustness of LLMs in high-risk scenarios involves `adversarially' attacking LLMs through the use of semantically equivalent adversarial rules \cite{ribeiro2018}, adversarial triggers \cite{wallace2019a} and even human-in-the-loop generation of adversarial examples \cite{wallace2019}.  Recent studies also explore the riskiness of certain LLM prompts \cite{halawi2023overthinking, shi2023large, turpin2023language}. While such research spotlights LLMs' vulnerabilities when adversarially prompted, we still expect LLMs to make judicious decisions about navigating a risk-reward tradeoff by being self-aware of when they might have higher probability of going wrong. Besides, although `risk' is frequently mentioned in adversarial attack research, as also in related literature on assessments of safety and bias \cite{ferrara2023should, zhou2022towards}, robustness-accuracy characteristics \cite{ko2023robustness}, and out-of-distribution (OOD) generalization \cite{grangier2022trade, xu2021raise}, a formalism is still lacking that identifies and defines multiple types of risk.  This paper proposes such a general formalism, including identifying two different risk categories that are relevant to LLMs, and novel metrics for measuring risk.

Initial research on LLMs' `self-understanding' of their own uncertainty, especially in the deep learning literature,  has predominantly relied on interpreting raw softmax probabilities of the final output layer as `confidence' scores \cite{vasudevan2019towards}. 
While studies such as \cite{guo2017calibration} have flagged these scores as potentially misleading and not \emph{genuinely} capturing the model's true uncertainty, \cite{kadavath2022language} highlighted that generative LLMs exhibit commendable calibration properties facing certain situations. These models can accurately predict which questions they will be able to answer correctly on diverse NLI tasks based on their confidence scores. Nevertheless, a quantitative assessment of risk, even for such models, has been lacking. 

\cite{jiang2021can} suggest that, when faced with uncertain situations,
LLMs can sometimes be poorly calibrated, with the confidence score estimation barely being correlated with the likelihood of the output being correct. Building on these observations, studies by \cite{jagannatha2020calibrating, jiang2021can, kuhn2022semantic} have endeavored to re-calibrate confidence, using mechanisms like entropy, or by crafting binary classifiers based on the given confidence scores. More recently, \cite{zhou2023navigating} focus on generative models, probing linguistic hedge markers in the models' outputs to evaluate their ability to discern uncertain situations. A further strand of research, such as by \cite{yin-etal-2023-large}, proposes benchmarks aiming to spotlight areas of knowledge where LLMs grapple with uncertainty. Concurrently, efforts by \cite{collins2023human} seek to minimize uncertainty in human-AI contexts by addressing risks originating from human errors.

In our work, we propose a novel re-adjusted confidence calibration for both generative and discriminative LLMs, aimed at discerning uncertain situations. We utilize a binary classifier calibration, previously shown to have exceptional performance \cite{kamath2020selective,jiang2021can}, as an experimental baseline. Rather than presenting a new benchmark, we devise a suite of risk injection functions designed to emulate `unanswerable' uncertain inference scenarios, and that can be applied to any existing multi-choice NLI benchmarks.

\section{Risk-adjusted calibration  framework}\label{Sec:formalism}

Before delving into the details of the proposed risk-adjusted calibration framework, we begin by laying out some basic terminology. First, we denote an NLI benchmark $I$ as consisting of a set of multi-choice inference \emph{instances} $i$, where each instance $i=(q, Y)$ consists of a \emph{prompt} $q$ and a finite set $Y$ of candidate choices. A prompt may be a question but not necessarily so. For example, instances in the aNLI benchmark have a \emph{observation} as prompt, with the candidate choices representing two hypotheses, one of which is the most likely explanation for the observation. The size of $Y$ (called \emph{choice cardinality}), or the number of candidate choices, is fixed for every instance in $I$ and denoted as $N$. 

Note that, for an instance $i=(q, Y)$, we do \emph{not} assume that there is always a correct answer in $Y$. We refer to $i$ as \emph{ambiguous} if it is not associated with a `ground-truth' answer $\hat{y} \in Y$. For ease of terminology, we use an indicator $\hat{i}$, with $\hat{i}=1$ denoting that the instance has a ground-truth answer (or is unambiguous), and $\hat{i}=0$ denoting ambiguity. 

The purpose of NLI is to predict the correct answer for as many instances as possible. Recent NLI systems, such as discriminative LLMs, make such predictions by outputting a real-value confidence score $c$ for each $y \in Y$. A common interpretation of such a score is that it is the confidence of the LLM in the corresponding candidate choice being correct. We define the \emph{confidence set} $C = \{c_1, c_2, ...c_{N}\}$ as the set of confidence scores for all candidate choices in an instance $i$, with the additional requirement that the scores sum to 1. 

Implicit in an NLI system is a \emph{decision rule}, denoted as $dr$, on the basis of which the system must decide whether to make an actual prediction $\hat{y}' \in Y$. In most systems, the default assumption is that the system always decides (i.e., $dr=1$) to make the prediction. Once it has decided to make a prediction, a \emph{selection rule} $sr$ is invoked. A typical selection rule, and one that we assume in this paper, is to predict the $\hat{y}'$ with the highest confidence in $C$ (i.e., $sr(C)=\hat{y}'$) with ties broken arbitrarily. However, other selection rules are also possible. 

Although the selection rule above is reasonable for appropriately calibrated confidence scores, the assumption that $dr$ is always 1 is \emph{risk-ignorant}. Intuitively, $dr$ should also depend on $C$ and its distribution. To take an extreme example, if $C$ is uniformly distributed, it is risky to decide, regardless of $sr$. Similarly, for ambiguous instances, $dr$ should ideally be 0. Next, we describe specific types of risk that can arise due to sub-optimal decision rules.

\subsection{Definitions and Formalism of Risk}

Using the terminology introduced earlier, we start by  defining decision risk:
\begin{definition}[Decision Risk] 
\label{def: decisionRisk}
Given an instance $i=(q,Y)$ and confidence set $C$ over $Y$, the \emph{decision risk} $r_d$ is set to 1 (and is otherwise 0) \emph{iff} at least one of two conditions is met: (1) the instance is unambiguous ($\hat{i} =1$) but $dr(C) = 0$; and (2) when the instance is ambiguous ($\hat{i} = 0$) but $dr(C) = 1$.
\end{definition}

Decision risk can be an important concern in many human-facing and otherwise critical applications, such as clinical decision-making, and geo-political forecasting. In such domains, ambiguity is not uncommon for a variety of complex reasons, including expert disagreement, and evolving situational knowledge. Note that, in some applications, one of the two conditions in the definition might be more consequential than the other; however, in this paper, we treat both equally as decision risks. 

Unfortunately, most existing NLI benchmarks typically treat all instances as equally unambiguous, with a definitive correct answer specified for each prompt. Therefore, to evaluate different decision rules, we introduce the notion of \emph{risk injection function} (RIF), denoted as $f$. When applied to an unambiguous instance $i=(q, Y)$, $f(i)$ yields a new instance $i'$ that is ambiguous. We refer to $i'$ as a \emph{risk-injected instance}, and the original unambiguous instance $i$ as a \emph{risk-free instance}. For example, $f$ may replace $q$ with an empty string, resulting in an instance with no theoretically correct `answer' by definition, since there is no prompt. In \emph{Experiments}, we describe other reasonable RIFs that can be applied to any multi-choice NLI benchmark to construct training and evaluation sets for decision rules.

\begin{figure*}[htbp]
  \centering
  \includegraphics[height=3.1in]{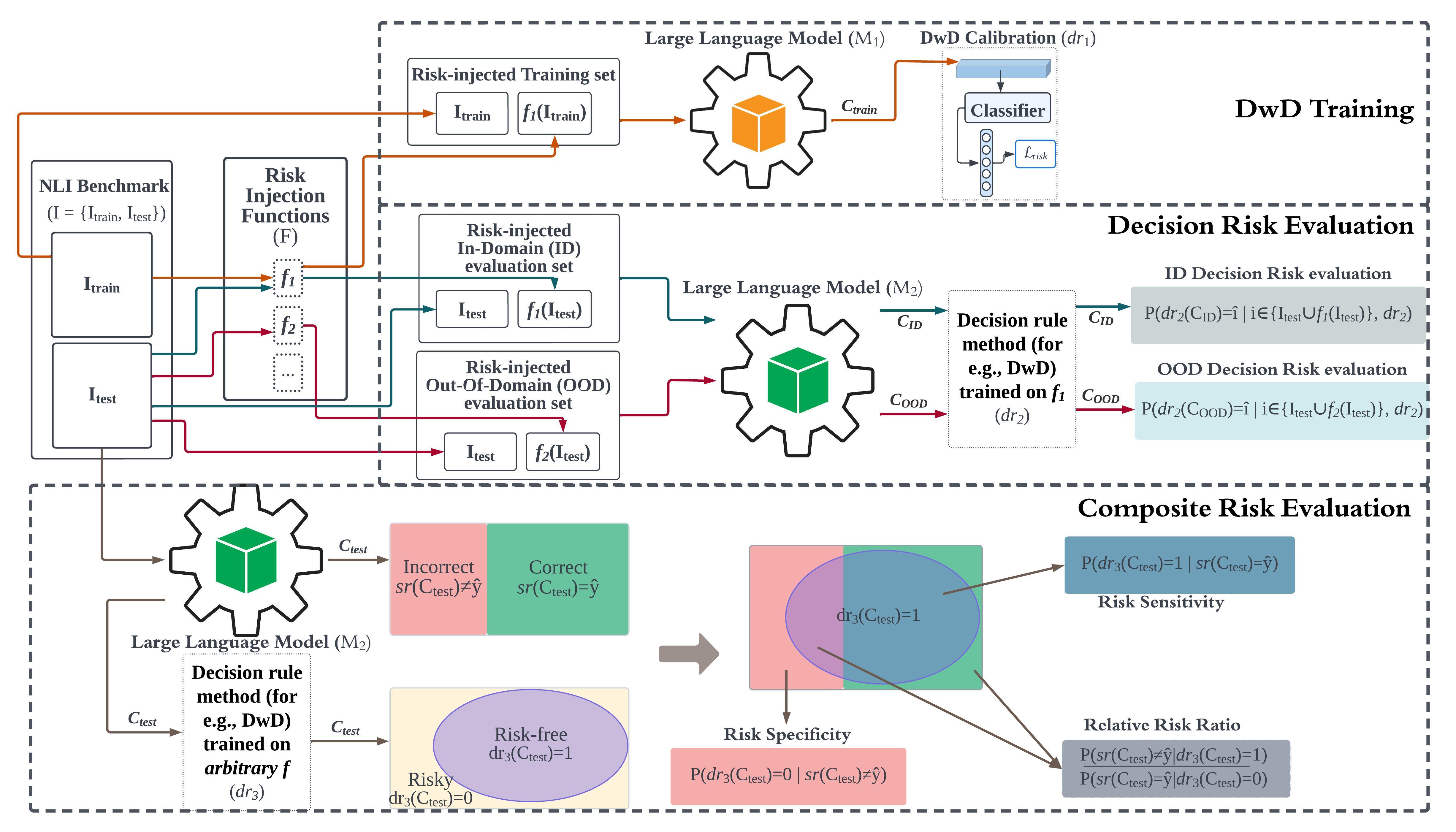}
  \caption{The proposed framework for measuring the risk-reward tradeoff of an LLM when it is embedded in a large inference system. The top pane presents the training process of our proposed decision rule, a risk-adjusted calibration named the \emph{DwD} method. The middle and last pane respectively depict the decision risk and composite risk evaluation of a large language model when influenced by a decision rule. Note that the large language model in the middle and bottom panes can differ from that in the top pane (but not necessarily so). Symbols are further described in the main text.}
  \label{fig:framework}

\end{figure*}

By construction, a well-performing decision rule $dr$ should output 0 for a risk-injected instance, and 1 for a risk-free instance. To evaluate the performance of $dr$, we can calculate the proportion of instances on which the decision rule yielded $dr(C)=\hat{i}$, the inverse proportion of the decision risk, when certain RIF(s) were applied to generate $i'$ in the evaluation set. We evaluate the robustness of $dr$ in two different scenarios: \emph{In-domain (ID)} and \emph{Out-of-domain (OOD)}. In an ID evaluation, the same RIF applied for training a decision rule is re-applied to an evaluation set containing unseen instances. Hence, the decision rule method has an opportunity to `learn' the RIF from the training data. In contrast, the OOD evaluation, meant to be harder and more realistic, uses different RIF(s) than the ones used during training; hence, the decision rule has no knowledge of these RIFs during the learning phase. The evaluation of decision risk  is illustrated in the middle pane in Figure \ref{fig:framework}.

Aside from decision risk, we can also define the \emph{composite risk}. This risk is motivated by the common situation where all instances are technically unambiguous, but where an underlying LLM finds some instances riskier (more likely to get wrong) than others. 

\begin{definition}[Composite Risk] 
\label{def: compositeRisk}
Given an instance $i = (q, Y)$ with a unique correct answer $\hat{y} \in Y$ and a confidence set $C$ over $Y$, the \emph{composite risk} $r_s$ is set to 1 (and is otherwise 0) \emph{iff} at least one of two conditions is met: (1) $dr(C)=0$ but the selection rule of the model yields a correct prediction $\hat{y}'=\hat{y}$; and (2) $dr(C)=1$ but the selection rule makes an incorrect prediction $\hat{y}'\neq \hat{y}$.
\end{definition}

Unlike decision risk, where the selection rule is assumed to be fixed and optimal, the composite risk depends on both the decision risk and selection risk. To capture this complexity, we define two metrics, \emph{risk specificity} and \emph{risk sensitivity}, inspired by similar concerns noted in communities like epidemiology \cite{runyan1998prevalence}. Risk specificity is computed as $P(dr(C)=0|sr(C)\neq \hat{y})$, while risk sensitivity is computed as $P(dr(C)=1|sr(C)= \hat{y})$. A low value for either suggests a specific type of composite risk, e.g., making decisions too aggressively than is warranted by the accuracy of the selection rule (or the underlying LLM being used) would lead to low risk specificity. 

To describe the trade-off between answering more questions and decreasing composite risks, we also use the \emph{relative risk ratio} (RRR) metric \cite{tenny2022relative} to evaluate $dr$. RRR is computed using the formula $\frac{P(sr(C)\neq \hat{y}|dr(C)=1)}{P(sr(C)= \hat{y}|dr(C)=0)}$. Although it measures composite risk of $dr$, its condition is reversed compared to risk sensitivity and specificity, defined earlier. When $RRR$ is significantly smaller than 1 (e.g., at the 95\% confidence level), it implies that $dr$ significantly reduces composite risk when it is 1, compared to when it abstains ($dr=0$). We schematize the evaluation of composite risk in the last pane of Figure \ref{fig:framework}, as it is important to the overall framework.

\subsection{Risk-adjusted Calibration Approach}

In the experiments, we employ both discriminative and generative models as inference systems. While both models can generate a confidence distribution for each inference instance, there is an inherent challenge: the absence of explicit decision rules within these models. Discriminative models naturally lack integrated decision rule mechanisms, and modern generative models, such as ChatGPT, have not disclosed their decision-making protocols, making any decision rules within them unpredictable from a user's standpoint. A decision rule that is compatible with both discriminative and generative LLMs, and that is somewhat independent of the LLM itself (and hence, generalizable), is clearly motivated.  An ideal decision rule should use all available information, such as the instance prompts, as well as the outputs from the underlying LLMs, to minimize the risks defined earlier.  

As noted in \emph{Related Work}, prior research introduced basic \emph{re-calibration} of confidence scores to better model an LLM's underlying uncertainty. We expand on this idea significantly by proposing a novel \emph{risk-adjusted} calibration method called\footnote{Shorthand for `Deciding when to decide'.} \emph{DwD} that can act as a \emph{decision rule} to help minimize the decision and composite risks of LLMs, especially in high-risk inference scenarios. DwD addresses the aforementioned challenges and does not have strong dependencies on the underlying LLMs, enabling it to be suitable for both discriminative and generative LLMs. 

While DwD is modeled as a binary classifier using a conventional machine learning algorithm,  it is differentiated in two aspects:

{\bf (i) RIF-based Training:} DwD uses available \emph{Risk Injection Functions} (RIFs) on NLI benchmarks to construct its training set, in contrast with other calibrators that do not take risk explicitly into account during the training. These RIFs expose the model to various scenarios where risks occur, ensuring that the method is well-equipped to handle real-world challenges. In \emph{Results}, we present the performance of DwD when trained using different RIFs, to ascertain which RIFs enhance its effectiveness, even in OOD scenarios.

{\bf (ii) Calibration Refinement with Risk Adjustment:} Previous work, such as \cite{kamath2020selective, jiang2021can}, has utilized the confidence distribution from inference systems as an initial calibration to form the decision rule. \emph{DwD} builds upon this foundation by refining the calibration. We use several features: the prompt length (in terms of the number of characters), the length of the predicted answer generated by the LLM, the standard deviation of the confidences across candidate choices per instance, the sentence embedding\footnote{Obtained using the pre-trained \emph{nq-distilbert-base-v1} model in \emph{SentenceTransformer}\cite{reimers2019sentence}.}  similarity between the prompt and each candidate choice, and the standard deviation of these embedding similarities. By using a diverse set of features, our aim is to help DwD be as robust as possible so that it is able to navigate composite risk better, and utilizes RIF training to maximum advantage. With these features, the \emph{DwD} method is trained using a random forest classifier with equal numbers of original and risk-injected instances (following the application of one RIF) as the training set. The probability of an inference instance being labeled as positive is interpreted as the confidence of DwD asserting $dr(C)=1$, which is used to estimate both the decision and composite risks. Note that the \emph{DwD} approach is specifically trained on discriminative LLM's confidence distribution, as generative LLMs such as GPT-3.5-turbo can only provide a rough confidence estimate. In \emph{Results} Section, we show that though the approach was exclusively trained on discriminative LLM's confidence distribution, it is adaptable to generative LLMs.

\section{Experiments}
\subsection{Datasets}
We use four established natural language inference (NLI) benchmarks (Abductive Natural Language Inference (aNLI) \cite{anli}, HellaSwag \cite{hellaswag}, Physical Interaction QA (PIQA) \cite{piqa} and Social Interaction QA (SocialIQA) \cite{socialiqa}) that are modeled as multiple-choice benchmarks, with exactly one choice being the correct response for a given `prompt'. For the validation sets of these benchmarks, ground-truth `answer keys' are publicly available and are used for the experiments. More details on format and performance are included in a supplementary Appendix. 

\subsection{Language Models}
We evaluate the generality of the proposed framework using two highly established language models: 
\subsubsection{RoBERTa Ensemble}
Prior to the release of generative LLMs, fine-tuned BERT-based variants were still the state of the art on many NLI tasks. To ease replication, we use the publicly available RoBERTa Ensemble model \cite{robertaensemble} which demonstrates effective fine-tuning capabilities on modest, task-specific datasets like those used for multi-choice NLI. To tailor the discriminative model to a particular NLI benchmark such as PIQA, we first fine-tune it on the benchmark's training set. We then apply the fine-tuned model to the evaluation set to assess performance. For each multiple-choice inference task, the model produces the \emph{confidence set} $C$ displaying the system's confidence in each potential answer.

\subsubsection{ChatGPT}
Given the remarkable accomplishments of recent generative large language models (LLMs), we use gpt-3.5-turbo\footnote{Accessed using OpenAI’s API: \url{https://openai.com/product}.} as an additional LLM to explore its interaction with the decision rule. Note that in our experiment, we directly combine the pre-trained GPT-3.5-turbo with the risk indicator trained by RoBERTa-related features. This setting allowed us to investigate how these adjusted risk scores interact with an (a priori) unknown model. In the experiments, the gpt-3.5-turbo model was prompted with a specific template to produce a confidence set $C$ (per instance) similar to the RoBERTa model described earlier. Details on the prompting mechanism, and the actual template used, are provided in the Appendix.
\subsection{Decision Rule Baselines and Selection Rule}
To evaluate the proposed DwD approach, we use three decision rule baseline methods in the experiments. Like DwD, all baselines described below treat the LLMs as black boxes and do not require access to the model's internal representations. Rather, for each instance, they only need the output (the confidence set $C$) of the model. 

{\bf Random Baseline.} Given an instance (into which risk may or may not have been injected), this baseline ignores the confidence set $C$ and randomly chooses between 1 (risk-free) and 0 (risky) with equal probability. 
Hence, if risk has been injected in half the instances in the evaluation set, the accuracy of its prediction will always be 0.5 (in expectation), and is intended to serve as a useful reference for evaluating more advanced decision rule methods.

{\bf ConfStd Baseline.} Inspired by \emph{MaxProb} \cite{hendrycks2016baseline}, this baseline uses the \emph{standard deviation} among all candidate choices' confidences (referred to as \emph{ConfStd}) to make its decision\footnote{Using MaxProb directly was found to yield near-random performance when evaluating decision risk (see Appendix).}. A lower \emph{ConfStd} corresponds to a higher risk of answering incorrectly. We use the training set of a benchmark to determine the optimal \emph{ConfStd} threshold below which a decision of 0 (risky) would be returned. This threshold is then applied during evaluation. 

{\bf Calibrator Baseline.} Finally, we used another baseline that relies on training a \emph{calibrator} \cite{dong2018confidence, kamath2020selective}. We opted to use the random forest model as the binary classifier, aligning with the consistency of our proposed \emph{DwD} method. Inspired by the experimental settings in \cite{kamath2020selective}, we designed the calibrator baseline using the prompt length (in terms of the number of characters), predicted answer ($\hat{y}$) length, and each candidate choice's confidence, as features. 

\paragraph{Selection Rule}
We consider only the standard confidence-based selection rule in this paper: given the confidence set $C$, we select $\hat{y}'$ as the selected choice, where $\hat{y}'$ is the choice assigned the highest confidence in $C$. Ties are broken arbitrarily.

\subsection{Risk Injection Functions}
The evaluation of in-domain and out-of-domain decision risk relies on a library of risk injection functions (RIFs). We designed three reasonable RIFs that can be used to inject risk into any instance $i=(q,Y)$ from an NLI benchmark $I$. Two of the RIFs operate at the level of the prompt (i.e., modify the prompt $q$, while leaving $Y$ intact), while one modifies the candidate choice-set $Y$: \\
{\bf No-Question RIF (NQ):} Replace the prompt $q$ with the empty string; the choice-set $Y$ remains as is;\\ {\bf Wrong-Question RIF (WQ):} Retain the candidate choice set $Y$  but replace the original prompt $q$ with a new prompt $q'$ from an unrelated instance in the same benchmark; \\{\bf No-Right-Answer RIF (NRA):} Retain the prompt $q$ and all \emph{incorrect} choices $Y-\hat{y}$ in the new candidate choice set $Y'$; also, add to $Y'$ an choice from another unrelated instance in the same benchmark\footnote{In contrast with NQ and WQ, this function modifies the choice-set $Y$, rather than $q$. Illustrative examples, code, and data for all RIFs are reproduced in the Appendix.}.



Note that these RIFs can be applied to both the training and validation set of a benchmark. For the purposes of training decision rule baselines (except \emph{Random}), risk-injected instances are labeled with 0, and original instances with 1. For conducting in-domain (ID) evaluations, we use each of these three RIFs in turn during both training and evaluation, and report results separately for each RIF. When conducting out-of-domain (OOD) evaluations, however, two sets of outputs are obtained for each RIF used during training. We average these two outputs when reporting OOD results\footnote{There are also other feasible ways of reporting out-of-domain results, other than a simple average, but the core conclusions do not change. We report on some alternatives in the Appendix.}. For example, if NQ is used as the training RIF, we would show the average of results for WQ and NRA (applied during evaluation).

\section{Results}

\begin{table*}[ht]
\centering \footnotesize
\begin{tabular}{llllll}
\hline
               NLI Benchmark            & $f$ & Random        & ConfStd             & Calibrator          & DwD                 \\ \hline
\multirow{3}{*}{PIQA}      & NQ       & 0.518 / 0.499 & 0.530* / 0.485      & 0.950*** / 0.500    & \textbf{0.970***} / \textbf{0.510}    \\
                           & WQ       & 0.499 / 0.518 & 0.550*** / 0.475    & 0.560** / 0.535     & \textbf{0.830***} / \textbf{0.680***} \\
                           & NRA      & 0.506 / 0.486 & 0.410 / 0.540***    & 0.540* / 0.505      & \textbf{0.840**} / \textbf{0.630***}  \\ \hline
\multirow{3}{*}{aNLI}      & NQ       & 0.503 / 0.509 & 0.590*** / \textbf{0.560***} & 1.000*** / 0.500    & \textbf{1.000***} / 0.500    \\
                           & WQ       & 0.479 / 0.492 & 0.640*** / 0.535    & 0.630*** / 0.520    & \textbf{0.940***} / \textbf{0.770***} \\
                           & NRA      & 0.483 / 0.504 & 0.460 / 0.630***    & 0.510 / 0.520       & \textbf{0.830***} / \textbf{0.885***} \\ \hline
\multirow{3}{*}{SocialIQA} & NQ       & 0.510 / 0.490 & 0.710*** / \textbf{0.575***} & 1.000*** / 0.500    & \textbf{1.000***} / 0.500    \\
                           & WQ       & 0.508 / 0.495 & 0.600*** / 0.580**  & 0.620* / 0.580*     & \textbf{0.790***} / \textbf{0.620*}   \\
                           & NRA      & 0.529 / 0.511 & 0.550*** / 0.655*** & 0.500 / 0.570**     & \textbf{0.630***} / \textbf{0.665***} \\ \hline
\multirow{3}{*}{HellaSwag} & NQ       & 0.489 / 0.503 & 0.580*** / \textbf{0.620***} & 1.000*** / 0.500    & \textbf{1.000***} / 0.500    \\
                           & WQ       & 0.502 / 0.501 & 0.610*** / 0.605*** & 0.730*** / 0.685*** & \textbf{0.940***} / \textbf{0.750***} \\
                           & NRA      & 0.502 / 0.501 & 0.630*** / 0.570*** & 0.660*** / 0.645*** & \textbf{0.820***} / \textbf{0.900***} \\ \hline
\end{tabular}
\caption{\footnotesize Accuracy of \emph{RoBERTa Ensemble} using a decision rule (Random / ConfStd / Calibrator / the proposed DwD method) when risky instances are present. $f$ indicates the RIF used during training, and performance is reported for both in-domain (ID) / out-of-domain (OOD) scenarios. The best performance is shown in bold. *, **, and *** indicate statistical significance with respect to \emph{Random} (in that row) at the 90\%, 95\%, and 99\% confidence level, respectively.}
\label{table: selective-pre-accuracy}
\end{table*}

\begin{table*}[h]
\centering \footnotesize
\begin{tabular}{llllll}
\hline
                         &                 & aNLI                         & HellaSwag                    & PIQA                         & SocialIQA                  \\ \hline
\multirow{7}{*}{RoBERTa} & Random          & 0.505/0.487/0.241            & 0.495/0.499/0.212            & 0.493/0.553/0.322            & 0.503/0.460/0.289          \\ \cline{2-6} 
                         & ConfStd$_{WQ}$     & 0.801/0.477/0.242            & 0.787/0.232/0.195         & 0.591/0.270/0.348            & 0.681/0.315/0.325          \\
                         & ConfStd$_{NRA}$    & 0.733/0.401/0.247            & 0.784/0.228/0.191          & 0.485/0.190/0.348            & 0.643/0.258/0.335          \\ \cline{2-6} 
                         & Calibrator$_{WQ}$  & 0.645/0.322/0.268          & 0.787/0.381/0.224         & 0.566/0.443/0.351            & 0.630/0.446/0.275          \\
                         & Calibrator$_{NRA}$ & 0.541/0.500/0.251            & 0.691/0.338/0.197          & 0.531/0.325/0.346            & 0.537/0.394/0.265          \\ \cline{2-6} 
                         & DwD$_{WQ}$         & \textbf{0.928}/\textbf{0.580}/\textbf{0.237} & \textbf{0.959}/\textbf{0.684}/\textbf{0.182}& \textbf{0.836}/\textbf{0.576}/0.308 & \textbf{0.781}/\textbf{0.531}/\textbf{0.257} \\
                         & DwD$_{NRA}$        & 0.818/0.560/0.247            & 0.843/0.569/0.193         & 0.811/0.568/0.342            & 0.599/0.519/0.279          \\ \hline
\multirow{2}{*}{ChatGPT} & DwD$_{WQ}$         & 0.912/0.120/0.503            & 0.954/0.054/0.560           & 0.826/0.206/\textbf{0.227}            & 0.758/0.231/0.405          \\
                         & DwD$_{NRA}$        & 0.846/0.160/0.248            & 0.727/0.170/0.203            & 0.824/0.188/0.342            & 0.516/0.444/0.273          \\ \hline
\end{tabular}
\caption{\footnotesize Sensitivity, specificity, and relative risk ratios (RRRs) (separated by `/')  of LLMs combined with different decision rule methods on the four original NLI benchmarks. All decision rule methods (except \emph{Random}) are always trained on RoBERTa-generated feature sets, even when testing ChatGPT as the LLM. Subscripts in Column 2 indicate the training RIF. All RRR results are significant with 95\% confidence.}
\label{table: sensitivity-specificity-odds}
\end{table*}

We begin this section by evaluating \emph{decision risks}, both in-domain (ID) and out-of-domain (OOD), encountered in NLI tasks when integrating different decision rules (DwD and the baselines described earlier) with \emph{RoBERTa Ensemble}.

\paragraph{Evaluating ID and OOD Decision Risks}

Table \ref{table: selective-pre-accuracy} reports the accuracy results (equivalent to 1$-$the proportion of decision risks) for \emph{RoBERTa Ensemble} incorporating different decision rule methods (DwD and the three baselines) on in-domain (ID) and out-of-domain (OOD) decision risk evaluation datasets. As earlier discussed, these datasets are built from the four NLI benchmarks by applying RIFs. 


In evaluating ID decision risk, results in Table \ref{table: selective-pre-accuracy} (first value of each cell) show that, across all benchmarks and settings, the accuracy of \emph{RoBERTa Ensemble}, when guided by the proposed DwD and Calibrator methods, outperformed its accuracy in conjunction with ConfStd and Random, by margins of 31.1\% and 15.3\%, respectively. This suggests the efficacy of learning-based methods (such as DwD and Calibrator), and of RIF-based training, in particular. In contrast, ConfStd, which directly utilizes RoBERTa's confidence as a decision rule, showed near-random performance for the PIQA benchmark, with slight improvements on other benchmarks. This finding aligns with the observation that the raw confidences of LLMs themselves tend to be poorly calibrated when confronted with uncertainty \cite{jiang2021can}.

 As the nature of decision risks confronted by LLMs is typically unknown, handling OOD decision risk is expected to be harder than handling ID risk.
Table \ref{table: selective-pre-accuracy} (second value in each cell) illustrates that \emph{RoBERTa Ensemble}, utilizing a DwD-generated decision rule trained with either WQ or NRA function, achieved the best performance with an average accuracy of 0.738. This outperformed other baselines by a substantial 18.9\% margin. Despite a slight decline of only 0.09 compared to its ID performance, the overall performance suggests that \emph{RoBERTa Ensemble} can be properly re-calibrated when exposed to both ID and OOD decision risks, when coupled with the correct decision rule method i.e., RoBERTa's raw confidence cannot directly be used as a reliable decision rule, but the DwD-adjusted decision rule can predict decision risk effectively.

ConfStd exhibits an insignificant decrease (0.002) in performance when evaluated on \emph{OOD} sets compared to its accuracy (0.571) on \emph{ID} decision risk evaluation sets. The result shows that, although ConfStd  has stable performance, its overall performance in distinguishing risky from risk-instances is low. Although prior studies, such as \cite{kadavath2022language}, suggested that LLMs can self-evaluate their assertions' validity and predict the accuracy of their answers based on confidence, these results also suggest that raw confidence should not be directly used as a decision rule.

\paragraph{Evaluating Composite Risks}
The efficacy of decision rules embedded in LLMs on composite risk evaluations is reported by sensitivity, specificity, and relative risk ratios (RRRs) in Table \ref{table: sensitivity-specificity-odds}. Note that the decision rules that were trained using the NQ RIF were found to be overfitting when evaluating decision risks, and also did not generalize; hence, the composite risk performance of LLMs conjoined with NQ-trained decision rules is not reported.

In Table \ref{table: sensitivity-specificity-odds}, \emph{RoBERTa Ensemble} with the DwD calibration (trained using the WQ RIF) was found to attain the highest sensitivity and specificity on all benchmarks. For `risk-free' instances, both ChatGPT and \emph{RoBERTa Ensemble} tend to yield confidence distribution with high `reference' values. When confronted with these instances and following the default \emph{selection rule}, models without a decision mechanism will directly produce correct predictions. Across the \emph{aNLI} and \emph{HellaSwag} benchmarks, the proposed DwD decision rule accurately guided the models on over 90\% of such instances. Compared to the decision rule baseline methods, DwD significantly reduces the language model's composite error, achieving an error reduction margin of 26.65\%. Note that these reductions are significant with 95\% confidence. Furthermore, on evaluation instances where the LLMs would have made incorrect predictions \emph{without} a decision rule, \emph{RoBERTa Ensemble}, when combined with the proposed DwD calibration, still achieves the highest performance, with an average specificity of 0.573, across four benchmarks. Compared to the average sensitivity performance of 0.822, a significant decline is observed, indicating that composite risk under the second condition poses a greater challenge.

The RRRs presented in Table \ref{table: sensitivity-specificity-odds} indicate that, under the guidance of decision rules (including the three baselines and the DwD method), \emph{RoBERTa Ensemble} and ChatGPT both display significantly reduced composite risks ($RRR<1$ at 95\% confidence level). 
When employing the DwD calibration (trained using \emph{WQ} RIF) as the decision rule for \emph{RoBERTa Ensemble}, the lowest RRR was obtained on almost all benchmarks, with PIQA being the sole exception. The average RRR of 0.246 indicates that the risk ratio when DwD prompts RoBERTa to make a prediction is around 25\% compared to the risk ratio when DwD prompts RoBERTa to skip the instances. Interestingly, when ChatGPT employs the DwD method (trained using the WQ RIF) as its decision rule, it exhibits lower RRRs on PIQA compared to RoBERTa using the same decision rule. This finding illustrates that even when the DwD method is exclusively trained on the confidence distribution of RoBERTa, it still exhibits a reasonable degree of generalization. This may enable it to offer plausible guidance to other LLMs, such as GPT-turbo-3.5, for confidently answering risk-free inference tasks. The finding is important because, unlike some of the more recent black-box LLMs, models like RoBERTa are more freely available in the open-source community and require (far) fewer computational resources to fine-tune. 

\begin{figure}[t]
  \centering
  \includegraphics[width=2.9in,height=2.25in]{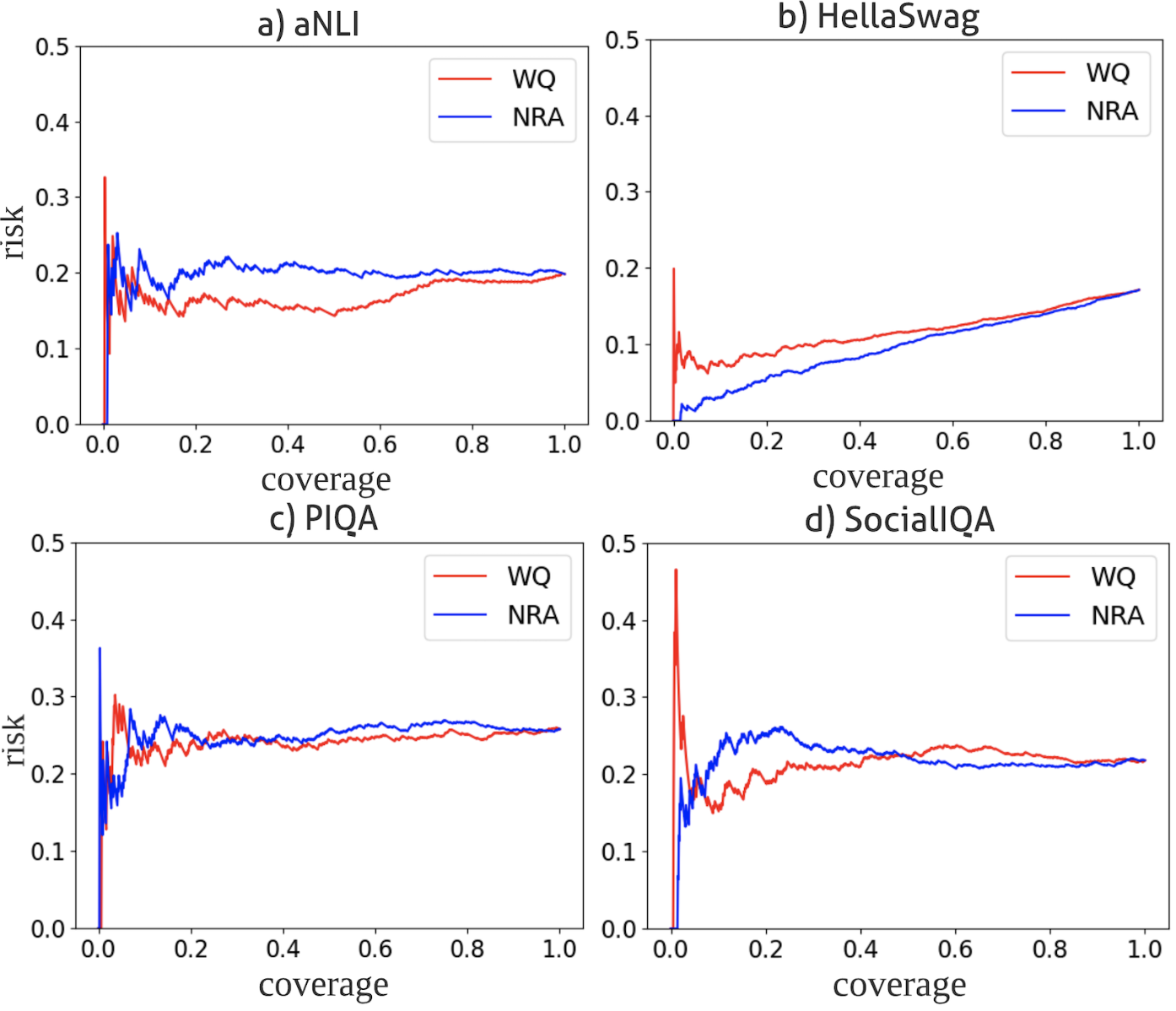}
  \caption{\footnotesize Risk-coverage curves for RoBERTa ensemble model that uses the proposed DwD method (WQ- and NRA-trained versions) as decision rule on all four benchmarks.}
  \label{fig:risk-coverage}
\end{figure}

\paragraph{Visualizing Risk-Coverage Tradeoff of DwD}\label{discussion}

We briefly discuss the risk-coverage tradeoff  of RoBERTa by plotting the risk-coverage curves (in Fig \ref{fig:risk-coverage}) for each benchmark, when the model was incorporated with the \emph{DwD-generated} decision rule. We expect a robust LLM with an effective risk-adjusted calibration to exhibit a lower aggregate risk when dealing with instances deemed as least risky, compared to riskier instances. Crucially, these risk-coverage curves suggest that the proposed decision rule can drastically decrease inference error across all four benchmarks. For example, with HellaSwag at 90\% coverage, the DwD calibration prompts the RoBERTa system to abstain from 1.8\% of incorrect instances, rising to 3.3\% at 85\% coverage. Considering the LLM's near-human performance, this improvement can significantly impact real-world applications, particularly given the cost-effectiveness of decision rule training. Results for NQ are not shown as it was found to be overfitting when evaluating composite risks, as discussed earlier, and its risk-coverage tradeoff is hence trivial.


%

\section{Conclusion}\label{conclusion}
This paper aimed to measure and improve the robustness of LLMs on inference tasks, using a risk-oriented evaluation framework that defined and focused on two types of risk: decision and composite. We also introduced risk-adjusted calibration method (DwD) that uses simple, but generalizable, training to adjust for these risks, and improve the robustness of the underlying LLM. DwD works with both generative and discriminative LLMs, and is agnostic to the details of the LLM itself. Detailed experiments, using four NLI benchmarks, demonstrate the practical utility of the proposed risk framework and the competitive performance of DwD.  

\newpage
\bibliography{aaai24}

\end{document}